%% file: root.tex
\documentclass[letterpaper, 10 pt, conference]{ieeeconf}
\IEEEoverridecommandlockouts
\usepackage{cite}
\usepackage{amsmath,amssymb,amsfonts}
\usepackage{stfloats}
\usepackage[caption=false,font=scriptsize]{subfig}
\usepackage{booktabs}
\usepackage{url}
\usepackage{hyperref}
\usepackage{cleveref}
\usepackage[nolist]{acronym}
\usepackage{tikz}
\usepackage{siunitx}
\usetikzlibrary{arrows.meta}
\usepackage{todonotes}
\usepackage{pgfplots}
\pgfplotsset{compat=1.18}

\definecolor{TUMBlue}{HTML}{6494eb}
\definecolor{TUMOrange}{RGB}{227, 114, 34}
\definecolor{myorange}{HTML}{E27121}
\definecolor{TUMGreen}{RGB}{73, 156, 0} 
\definecolor{TUMRed}{RGB}{255,0,0} 
\definecolor{TUMBlack}{RGB}{0,0,0} 
\definecolor{TUMPurple}{RGB}{160,32,240} 

\def\BibTeX{{\rm B\kern-.05em{\sc i\kern-.025em b}\kern-.08em
    T\kern-.1667em\lower.7ex\hbox{E}\kern-.125emX}}
\DeclareMathOperator\erf{erf}

\newcommand\copyrighttext{%
	\footnotesize This work has been submitted to the IEEE for possible publication. Copyright may be transferred without notice, after which this version may no longer be accessible.
}

\newcommand\copyrightnotice{%
	\begin{tikzpicture}[remember picture,overlay]
	\node[anchor=south,yshift=10pt, xshift=10pt] at (current page.south) {\fbox{\parbox{\dimexpr\textwidth-\fboxsep-\fboxrule\relax}{\copyrighttext}}};
	\end{tikzpicture}%
}
\begin{document}
\bstctlcite{BSTcontrol}
\title{Precise and Efficient Collision Prediction under \\ Uncertainty in Autonomous Driving}

\author{Marc Kaufeld, Johannes Betz 
\thanks{M. Kaufeld and J. Betz are with the Professorship of Autonomous Vehicle Systems, TUM School of Engineering and Design, Technical University of Munich, 85748 Garching, Germany; Munich Institute of Robotics and Machine Intelligence (MIRMI)}
}

\maketitle
\copyrightnotice
\begin{abstract}
This research introduces two efficient methods to estimate the collision risk of planned trajectories in autonomous driving under uncertain driving conditions. Deterministic collision checks of planned trajectories are often inaccurate or overly conservative, as noisy perception, localization errors, and uncertain predictions of other traffic participants introduce significant uncertainty into the planning process. This paper presents two semi-analytic methods to compute the collision probability of planned trajectories with arbitrary convex obstacles. The first approach evaluates the probability of spatial overlap between an autonomous vehicle and surrounding obstacles, while the second estimates the collision probability based on stochastic boundary crossings. Both formulations incorporate full state uncertainties, including position, orientation, and velocity, and achieve high accuracy at computational costs suitable for real-time planning. Simulation studies verify that the proposed methods closely match Monte Carlo results while providing significant runtime advantages, enabling their use in risk-aware trajectory planning.
The collision estimation methods are available as open-source software: \url{https://github.com/TUM-AVS/Collision-Probability-Estimation}
\end{abstract}
\begin{keywords}
Autonomous vehicles, trajectory planning, collision probability, risk estimation.
\end{keywords}
\input{text}

\bibliographystyle{IEEEtran}
\bibliography{literature.bib}

\begin{acronym}
\acro{av}[AV]{autonomous vehicle}
\acro{pdf}[PDF]{probability density function}
\acro{ttc}[TTC]{time-to-collision}
\acro{psd}[PSD]{proportion of stopping distance}
\acro{cog}[CoG] {center of gravity}
\end{acronym}


\end{document}

%% file: text.tex
\section{Introduction}
In autonomous driving, continuous collision checking is necessary to enable safe and collision-free motion planning~\cite{nieSurveyContinuousCollision2020}. In each planning cycle, planned trajectories must be checked against potential future collisions with surrounding obstacles \cite{annellProbabilisticCollisionEstimation2016}. 
Therefore, the \ac{av} needs to carefully perceive, track, and predict the behavior of vehicles in its proximity. 
However, noisy and incomplete sensor data and localization errors limit the accuracy of the environment model and own state estimation. Additionally, the future behavior of other traffic participants remains difficult to predict. These uncertainties propagate into the planning process, such that deterministic collision checks, e.g., based on hierarchical bounding volumes \cite{jimenezCollisionDetectionAlgorithms1998, pekCommonRoadDrivabilityChecker2020} are not reasonably applicable for online motion planning in autonomous driving.

Planning and control strategies must therefore assess the likelihood of collisions while taking into account prevailing uncertainties to minimize both the risks to the safety of surrounding traffic participants and risks to the vehicle’s own reliable operation \cite{schwartingPlanningDecisionMakingAutonomous2018, mcallisterConcreteProblemsAutonomous2017}. 
A principled way to achieve collision awareness is to incorporate risk metrics into decision-making, which allows the planner to explicitly trade off safety against performance. 
One possible strategy is to calculate the probability of collision along the planned trajectory as the probability of spatial overlap at each point in time \cite{lambertFastMonteCarlo2008}, or as the likelihood of colliding with an obstacle's boundary over a certain period of time \cite{altendorferNewApproachEstimate2021}.

\begin{figure}[t]
    \centering
\includegraphics[width=0.9\columnwidth,trim=2cm 3cm 2cm 3cm, clip,]{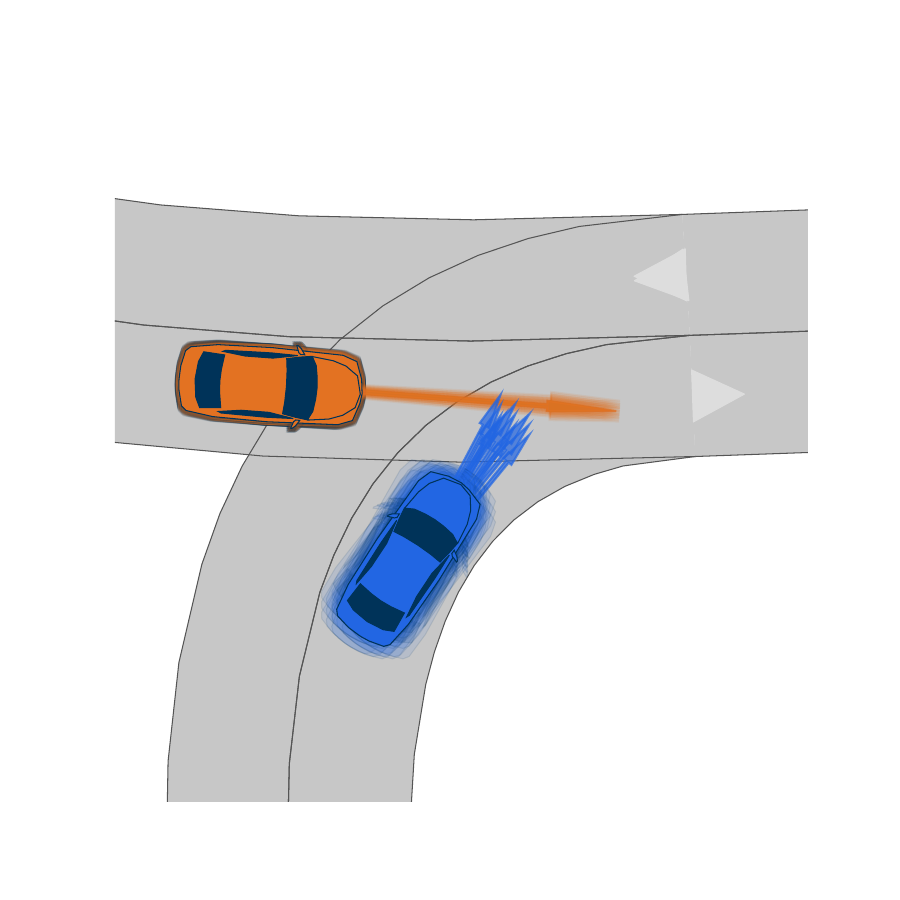}
    \caption{Exemplary scenario with large state uncertainties for the obstacle vehicle (blue) and small ones for the \ac{av}, illustrating the necessity of a probabilistic collision risk measure in motion planning.}
    \label{overview}
\end{figure}

This paper presents two methods to efficiently and accurately estimate the expected risk of planned trajectories for an \ac{av} under uncertain conditions. 
We propose two semi-analytic strategies to assess the collision probability based on a probabilistic prediction of surrounding obstacles, incorporating full state uncertainties. 
Our results indicate that both methods are capable of correctly forecasting the collision probability under appealing computation times.

\section{Related Work}
Usually, two paradigms for handling the risk of collisions along a planned trajectory can be  distinguished in autonomous driving \cite{akellaRiskAwareRoboticsTail2025}: 
Worst-case approaches safeguard against the most severe safety violations, achieving a high level of safety, but often result in overly conservative and impractical driving behaviors. 
Such formulations are often based on bounding volumes, e.g. spheres~\cite{xiaoyanResearchCollisionDetection2016}, ellipsoids~\cite{choiContinuousCollisionDetection2006} or oriented bounding boxes~\cite{gottschalkOBBTreeHierarchicalStructure1996},  with
safety margins around the obstacles to account for measurement errors~\cite{huNavigationControlMobile1991,zapataReactiveBehaviorsFast1994, pekCommonRoadDrivabilityChecker2020}. 
Reachability analysis is then performed to obtain collision-free driving corridors \cite{shkolnikReachabilityguidedSamplingPlanning2009, kousikBridgingGapSafety2020, manzingerUsingReachableSets2021, kochdumperRealTimeCapableDecision2024}. However, in the worst case, the \ac{av} is subject to the freezing robot problem, when no collision-free driving corridor can be found \cite{trautmanUnfreezingRobotNavigation2010}. 

Risk-aware techniques, in contrast, explicitly account for the distribution of possible outcomes, aiming to balance efficiency and safety. 
The objective of risk-aware planning is to enable autonomous vehicles to interact safely and intelligently with their environment while allowing a certain level of risk. 
Often, planned trajectories are evaluated based on criticality measures, such as \ac{ttc} or \ac{psd} \cite{janssonCollisionAvoidanceTheory2005, westhofenCriticalityMetricsAutomated2023}. However, these metrics assume a deterministic and simplified one-dimensional collision problem, and their results can differ significantly from the real values due to ignored uncertainties, leading to unexpected collisions \cite{lambertCollisionProbabilityAssessment2008}.
Other authors propose barrier functions centered around the midpoint of obstacles to incorporate risk awareness. They calculate a distance-dependent risk value while neglecting the shape of the obstacle \cite{mullerTimeCourseSensitiveCollision2020, akellaRiskAwareRoboticsTail2025}. These barrier functions can incorporate the positional uncertainty of the obstacles by defining a probability-dependent threshold value, but remain an estimated value\cite{geisslingerEthicalTrajectoryPlanning2023, schwartingSafeNonlinearTrajectory2018}.
An alternative approach is to calculate the probability of collision as the percentage of spatial overlap between the ego vehicle and an obstacle, whose position is assumed to be uncertain and can be described by a \ac{pdf}.
The collision probability is then included as a risk constraint in motion planning \cite{brudigamStochasticModelPredictive2023, batkovicSafeTrajectoryTracking2022}. Thereby, the position of the ego vehicle and the relative orientation between ego and obstacle are usually assumed to be deterministic.
The percentage of spatial overlap can generally be obtained using Monte Carlo Sampling \cite{lambertFastMonteCarlo2008, gouletProbabilisticConstraintTightening2022}, which is computationally expensive and therefore not applicable under real-time conditions in autonomous driving \cite{lambertCollisionProbabilityAssessment2008}.
As a result, the collision probability needs to be calculated more efficiently, e.g., by integrating the positional \ac{pdf} over the collision domain. However, for arbitrarily shaped objects, these integrals often lack an analytic solution.
Some authors therefore propose simplifications, e.g., by assuming robots approximated as points \cite{dutoitProbabilisticCollisionChecking2011, patilEstimatingProbabilityCollision2012}, (multiple) spheres \cite{parkPredictiveEvaluationShip2017, parkFastBoundedProbabilistic2018, tolksdorfFastCollisionProbability2024},  or rectangles with deterministic orientation \cite{philippAnalyticCollisionRisk2019}. 

All approaches so far assume statistical independence between states of an obstacle at different time steps. However, since this assumption does not hold for predicted trajectories, the collision probability of planned trajectories cannot be accumulated over time as a total collision risk \cite{schreinerBayesianEnvironmentRepresentation2016}.  
Consequently, a common approach for evaluating the collision probability along a trajectory or a certain time horizon is to use the maximum risk value over that period \cite{janssonCollisionAvoidanceTheory2005, janssonFrameworkAutomotiveApplication2008}.
Alternatively, in \cite{mullerTimeCourseSensitiveCollision2020}, a survival probability is proposed that uses the spatial overlap and utilizes a truncation approximation to model the temporal dependency of the collision risk.
The authors in \cite{altendorferNewApproachEstimate2021} derive an alternative method based on boundary crossing probabilities in stochastic vector processes; however, they do not consider orientation distributions or moving ego vehicles and only provide an approximate solution.

\subsection{Contributions}
In this paper, we address the precise prediction of the collision risk for future trajectories 
to enable \acp{av} to plan efficiently and safely in a risk-aware manner.
\begin{itemize}
    \item We derive two exact mathematical formulations to calculate the collision probability between arbitrary convex-shaped obstacles.
    \item Both methods are capable of coping with uncertain conditions prevalent in the \ac{av} software stack by incorporating uncertainties about the pose, orientation, and velocity of surrounding obstacles.
    \item Our approaches achieve compelling computation times, enabling their applicability in computing-time-limited applications in autonomous vehicles.
\end{itemize}

\section{Methodology}
In this section, we state both semi-analytic formulations for calculating the collision probability. First, we derive a method based on spatial overlap in \cref{ssec:PSO}, followed by a concept utilizing the probability of boundary crossings in \cref{ssec:BCP}.
We consider the situation where an ego \ac{av} is controlled by a sampling-based motion planner in a receding horizon technique \cite{kaufeld2025mprbfnlearningbasedvehiclemotion}.
In each time step $t$, the vehicle plans a  new trajectory $\mathbf{\zeta}_i(t)=\{(\mathbf{x}_{e}^\tau, \mathbf{v}_e^\tau), \,{\tau \in [t,t+T]}\}$ - represented as a sequence of states $\mathbf{x}_{e}^\tau = (x_{e}^\tau,y_e^\tau,\theta_e^\tau)$ and velocities $\mathbf{v}_e^\tau = (v_{x,e}^\tau,v_{y,e}^\tau)$
- by sampling and evaluating a set of potential trajectories with a predefined planning horizon~$T$. 
The states thereby consist of the coordinates at the vehicle center $(x_{e}^\tau,y_e^\tau)$ and the vehicle heading~$\theta_e^\tau$.
The trajectories are evaluated based on some cost terms, and the optimal one is followed for a time increment before recalculating a new set of trajectories in the next time step $t + \Delta t$. The planner thereby considers different cost terms related to smoothness, progress, and safety. The collision probability described in the following is thereby used to assess the safety of a planned trajectory. 
In order to calculate the likelihood of collision along a trajectory, we need to predict the pose of surrounding obstacles.
Perceiving the position and orientation of surrounding obstacles is subject to sensor errors and detection inaccuracies and is thus prone to uncertainty. These uncertainties are further amplified by the necessity to forecast the behavior of traffic participants, which introduces additional aleatoric and epistemic uncertainties from the prediction. 
Therefore, we assume the predicted states and velocities of obstacles to follow a (time-dependent) \ac{pdf}
$p_{o}(\tau) = p(\mathbf{x}_{o}^\tau, \mathbf{v}_e^\tau)$
. 
Similarly, the planned trajectory of the ego \ac{av} could be subject to uncertainties $p_{e}(\tau) = p(\mathbf{x}_{e}^\tau,\mathbf{v}_e^\tau)$, due to potentially not negligible localization or sensor errors as well as process noise. 
While the state and velocity uncertainties can be correlated, we assume that the \acp{pdf} of obstacles $p_o$ and the ego \ac{av} $p_e$ are independent from each other.

\subsection{Probability of Spatial Overlap}
\label{ssec:PSO}
Since the following calculations are valid for all time steps, we will omit the time index $\tau$ in the following chapter for better readability.
As the collision between the ego \ac{av} and a surrounding obstacle can be seen as the overlap of their occupancies, the probability of collision at any time step can be expressed as the probability of spatial overlap~\cite{altendorferNewApproachEstimate2021}.
After marginalizing out the velocity components, the pose distributions of the ego and obstacle are obtained as
\begin{equation}
    p_{\mathbf{x}_j} =\int p_{_j} \,d\mathbf{v}, \quad j\in \{o,e\}.
\end{equation}
The probability of spatial overlap can then be expressed as an integral over the joint pose distributions,
\begin{equation}
\label{eq:pc}
    P_{c}= \iint I_c(\mathbf{x}_{e},\mathbf{x}_{o})p_{\mathbf{x}_{e}}p_{\mathbf{x}_{o}} \,d\mathbf{x}_{o}\,d\mathbf{x}_{e},
\end{equation}
where the indicator function $I_c$ describes whether the occupied regions intersect, 
\begin{equation}
    I_c(\mathbf{x}_{e},\mathbf{x}_{o}) = \begin{cases}
        1 & \mathtt{Occ}(\mathbf{x}_{e}) \cap \mathtt{Occ}(\mathbf{x}_{o}) \neq \emptyset \\
        0 & \text{else.}
    \end{cases}
\end{equation}
Because the indicator function introduces a discontinuity, a direct analytic solution of \cref{eq:pc} is, in most cases, not possible.
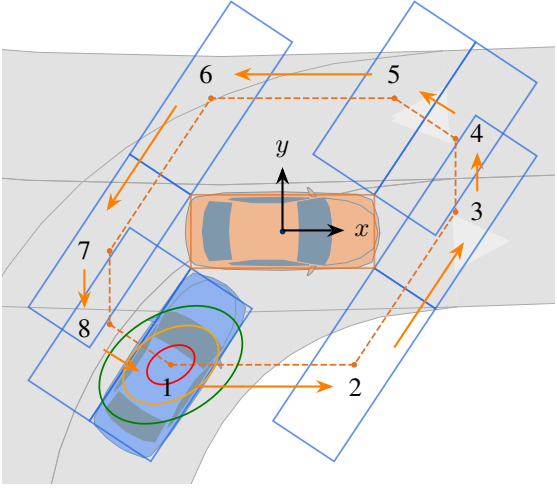
\begin{figure}[t]
    \centering
\input{figures/collision_octagon}
\caption{Combined collision volume of the ego \ac{av} (orange) and an obstacle vehicle (blue). Vehicles are approximated as rectangles. The ellipses over the blue vehicle illustrate the combined positional covariance $\mathbf{\Sigma_x}$ }
\label{fig:colloc}
\end{figure}
A more tractable formulation is obtained by considering the relative pose between the ego \ac{av} and an obstacle \linebreak $\mathbf{x}=\mathbf{x}_{o}- \mathbf{x}_{e}$. The probability density of $\mathbf{x}$ follows from the convolution of the two marginal pose densities, \linebreak $p_{\mathbf{x}} =p_{\mathbf{x}_e} * p_{\mathbf{x}_o}.$

When both uncertainties are modeled as multivariate normal distributions $p_i \sim \mathcal{N}\left(\mathbf{\mu}_i, \mathbf{\Sigma}_i\right), \; i \in \{\mathbf{x}_o,\mathbf{x}_e\}$, the convolution has a closed-form solution, which is again Gaussian,
\begin{align}
p_{\mathbf{x}} &\sim\mathcal{N}(\mathbf{\mu}_{\mathbf{x}}, \mathbf{\Sigma}_{\mathbf{x}}) \\
&\text{with}\nonumber \\
    &\mathbf{\mu}_{\mathbf{x}} = \mathbf{\mu}_{\mathbf{x}_o}-\mathbf{\mu}_{\mathbf{x}_e}, 
    &\mathbf{\Sigma}_{\mathbf{x}} = \mathbf{\Sigma}_{\mathbf{x}_o} + \mathbf{\Sigma}_{\mathbf{x}_e}.\nonumber
\end{align}
With this relative formulation, \cref{eq:pc} reduces to integrating $p_{\mathbf{x}}$ over a combined collision volume $\Omega$,
\begin{equation}
        P_{c}=\iiint\limits_{\Omega} p_{\mathbf{x}}(x,y,\theta) \,dx\,dy \,d\theta.
    \label{eq:p}
\end{equation}
The resulting value corresponds to the probability that the center of the obstacle lies within the collision volume spanned around the ego vehicle.
When neglecting the orientational variance, the collision volume is defined by the Minkowski sum of the shapes of the colliding obstacles $\Omega = S_e \oplus S_o$. 

In the two-dimensional representation, this is equivalent to the trace of the center of the obstacle swept around the hull of the ego \ac{av} as illustrated in \cref{fig:colloc}.
For any combination of two convex polygons, the collision volume is again a convex polygon and can be efficiently calculated by merging the edges based on their orientation with respect to the origin~\cite{cgal}. For collision checking in autonomous driving, where vehicles are usually represented as rectangles, the calculation of the Minkowski sum simplifies to adding up the oriented and signed half-length and -width of one vehicle to the shape of the other. 

\begin{figure}[t]
    \centering
\input{figures/polyhedron}
\caption{The orientation-dependent collision polyhedron. It is the same scenario as in  \cref{fig:colloc} but with an orientation variance of $\sigma_\theta^2 = \SI{0.01}{\square\radian} $. The third dimension illustrates the change in the shape of the collision volume. The distribution on the right indicates the \ac{pdf} of the relative orientation $\theta$.}
\label{fig:polyhedron}
\end{figure}
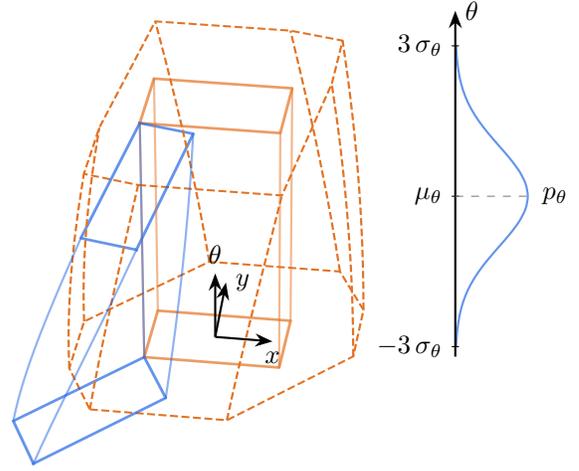

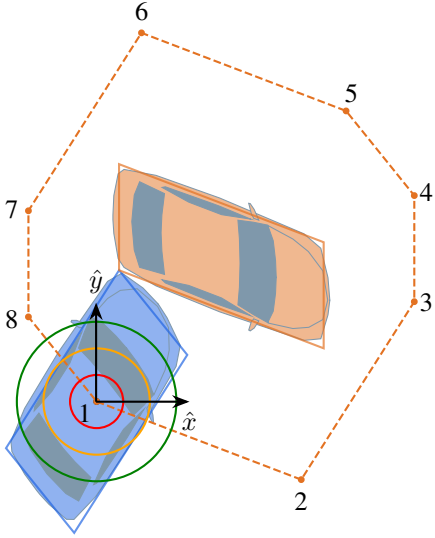
\begin{figure}[t]
    \centering
    \input{figures/whitened_oct}
    \caption{The collision volume from the scenario illustrated in \cref{fig:colloc} after whitening transformation.} 
    \label{fig:white}
\end{figure}

If the relative orientation between the ego \ac{av} and the obstacle is not deterministic, the collision volume is not fixed, but depends on the marginal orientation distribution, i.e. $\Omega = \Omega(p_\theta)$, resulting in a polyhedron as shown in \cref{fig:polyhedron}. 
With conditioning on the orientation, the integral \eqref{eq:p} can be rewritten as
\begin{equation}
        P_{c}=\int\limits_\theta \Bigl(\int\limits_{ y(\theta)} \int\limits_{ x(\theta)} p_\mathbf{x}( x,  y \mid \theta) \,dx\,dy\,\Bigr) p_\theta \,d\theta.
        \label{eq:int}
\end{equation}
So far, we have made no assumptions about the shape of the distribution. Since they might have correlated components and could be elongated, e.g., through rotated anisotropic distributions, we perform Cholesky whitening~\cite{Kessy2018} to transform the relative pose and the collision volume to be isotropic and centered around the center of the obstacle,

\begin{align}
    &\hat{\mathbf{x}} =\mathbf{W_x}(\mathbf{x}-\mathbf{\mu}_\mathbf{x}),\\
     &\mathbf{\hat{\Sigma}_x}= \mathbf{W_x}\mathbf{\Sigma_\mathbf{x}} \mathbf{W_x}^T = \mathbf{I}.
    \label{eq:white}
\end{align}
$\mathbf{W}$ is a transformation matrix such that $\mathbf{W_x}^T\mathbf{W_x}=\mathbf{\Sigma}^{-1}$ and can, for example, be obtained by Cholesky decomposition of the inverse covariance  $\mathbf{\Sigma_\mathbf{x}}^{-1} = \mathbf{LL}^T$ as $\mathbf{W}=\mathbf{L}^T$.
By whitening, all directions become independent and share the same uniform variance $\mu_{i}=0, \sigma_i=1, \; i \in \{\hat x,\hat y,\hat \theta \} $ (compare \cref{fig:white}), making the solution procedure numerically more stable and efficient without changing the final result. 
For a fixed $\theta$  and a convex and whitened collision volume with $N$ sides, the inner integral in \cref{eq:int} can now be written as 
\begin{align}
            P_{c}(\mathbf{\hat x}\mid \hat \theta)&= \sum_{n=1}^N\!\int\limits_{\hat x= x_{l,n}(\hat \theta)}^{x_{h,n}(\hat \theta)}\!\!\!\!\!p_{\hat x}(\hat x)\Bigl(\!\!\!\int\limits_{\hat y=0}^{y_{h,n}(\hat x, \hat \theta)}\!\!\!\!\!\! p_{\hat y}(\hat y)   \,d \hat y \Bigr)\,d\hat x ,
\end{align}
where the sequential integration limits $x_{l,n},x_{h,n}$ are the $\hat x$-coordinates of the $N$ corners of the collision volume at a given relative orientation $\hat \theta$ in anti-clockwise direction
and $\hat y_{h,n}$ is defined by their connecting line $\hat y_{h,n}(\hat x, \hat \theta)=m_n(\hat \theta)\hat x +b_n(\hat \theta)$ (see \cref{fig:white}).
Evaluating the inner integral, the formulation reduces to one dimension, 
\begin{align}
    &P_{c}(\mathbf{\hat x}\mid \hat \theta) =\! \sum_{n=1}^N\!\int\limits_{\hat x=x_{l,n}(\hat \theta)}^{x_{h,n}(\hat \theta)}\!\!\!\!\!\!p_{\hat x}(\hat x)\biggl(F_{\hat y}(y_{h,n})\!-\!F_{\hat y}(0)\biggr)\,d \hat x,
\end{align}
with $F_{\hat y}$ being the cumulative distribution function of $p_{\hat y}$.
This expression can be solved numerically very efficiently.
Assuming a multivariate Gaussian distribution, the probability of spatial overlap for one time step becomes
\begin{align}
    P_{c}(\mathbf{\hat x}\mid \hat \theta)& = \sum_{n=1}^N \frac{1}{\sqrt{8\pi}}\!\!\!\!\!\!\!\!\int\limits_{\hat x=x_{l,n}(\hat \theta)}^{x_{h,n}(\hat \theta)}\!\!\!\!\!\!\! \exp\hspace{-.05cm}\biggl(\!\!\frac{-\hat x^{2}\!}{2}\biggr)\erf\hspace{-.03cm}\biggl(\!\frac{y_{h,n}}{\sqrt{2}}\!\biggr)\, \,d\hat x, \\
    &\text{with} \nonumber \\
    &y_{h,n} = m_n(\hat \theta)\hat x +b_n(\hat \theta).\nonumber
\end{align}
For additional consideration of the orientation uncertainty, one now needs to solve the outer integral in  \cref{eq:int},
\begin{align}
P_{c} = \int\limits_{\hat \theta}p_{\hat \theta}(\hat \theta) P_{c}(\mathbf{\hat x}\mid \hat \theta) \,d\hat \theta,
\label{eq:p_ov}
\end{align}
which again can be solved effectively by numerical integration. This expression results in the probability of partial overlap at a certain time step under position and orientation uncertainties.
To evaluate the collision risk along a trajectory, one can either take the maximum value  
    $P_c = \max_{\tau\in T} \{P_{c}^\tau\}$,
or, if temporal independence of the states is assumed, the total collision probability over a trajectory with a planning horizon of $T$ is 
\begin{equation}
    P_c = 1-\prod_{\tau=0}^T (1-P_{c}^\tau).
    \label{eq:ind}
\end{equation}

\subsection{Boundary Crossing Probability}
\label{ssec:BCP}
In the previous chapter, we presented a solution method for the collision probability between an ego \ac{av} and an obstacle by estimating the spatial overlap. However, real-world collisions are better characterized by vehicles crashing into each other and thus crushing the chassis of the other vehicles from the front, rear, or sides. 
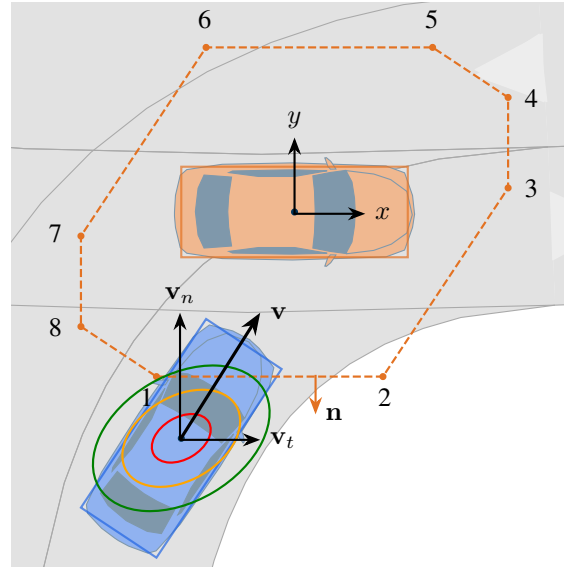
\begin{figure}[t]
    \centering
    \vspace*{2mm}
 \input{figures/boundary_cross}
    \caption{Illustration of the normal velocity for the boundary crossing rate along the segment $1-2$ of the collision volume.}
    \label{fig:cep}
\end{figure}
To reflect this behavior more accurately, we present a formulation for the calculation of the collision probability based on stochastic boundary crossing processes as proposed in 
\cite{altendorferNewApproachEstimate2021}.
The number of boundary crossings of a particle within a time interval corresponds to the integration of the boundary crossing rate over this interval.
Analogously, the probability of boundary crossings is defined as the time integral over the crossing probability rate $\dot P_{bc}$,
\begin{equation}
    P_c = \int_{\tau=t}^T  \dot P_{bc} \, d\tau.
    \label{eq:int_bc}
\end{equation}
For collision checking, the probability rate of boundary crossings is derived in detail in \cite{altendorferNewApproachEstimate2021}. With a state and velocity distribution $p(\mathbf{x,v})$ at the center point of an obstacle, it is defined as the expected velocity pointing into the collision volume, while the center is located at the boundary of the collision volume.
The collision probability rate is then defined as 
\begin{align}
    \dot P_{bc} =
    & - \!\!\!\iint\limits_{\substack{\mathbf{x}  \in \partial \Omega,\\ v_n\leq0}}v_n p( \mathbf{x},\mathbf{v})\,d \mathbf{x}\,d \mathbf{v},
    \label{eq:cep}
\end{align}
with $v_n = \mathbf{n}^T\!(\mathbf{x})\mathbf{v}$  denoting the velocity component normal to the volume boundary $\partial \Omega$.
By making use of the conditional distribution of the velocity and after marginalizing the tangential velocity, we can rewrite \cref{eq:cep} as 
\begin{align}
    \dot P_{bc} = & - \!\!\!\!\!\int\limits_{\mathbf{x}  \in \partial \Omega}\biggl(\!\!\!\int\limits_{ v_n\leq0}v_n p(v_n\mid \mathbf{x})\,d v_n \biggr)p_{\mathbf{x}}(\mathbf{x})\,d \mathbf{x} \nonumber
     \\
    &= \!\!\!\!\!\int\limits_{\mathbf{x}  \in \partial \Omega}\biggl(
\mathbb{E}\left[v_n^- \mid \mathbf{x}\right]\biggr)
\; p_{\mathbf{x}}(\mathbf{x}) \, d\mathbf{x}.
\end{align}
\begin{figure*}[!t]
\centering
\subfloat[][Scenario 1 without orientation uncertainty: Obstacle vehicle approaches the ego \ac{av} from the front left]{\label{fig:ss1}
\input{figures/scene1}}
\hfill
\subfloat[][Scenario 1 with orientation uncertainty: The different collision boundaries indicate the uncertainties in the relative orientation]{\label{fig:ss11}
\input{figures/scene1_orient}}
\hfill
\subfloat[][Scenario 2 without orientation uncertainty: Obstacle vehicle approaches the ego \ac{av} with offset from the side]{\label{fig:ss2}
\input{figures/scene2}} 
\hfill
\subfloat[][Scenario 2 with orientation uncertainty: The different collision boundaries indicate the uncertainties in the relative orientation]{\label{fig:ss22}
\input{figures/scene2_orient}} 
\caption{Considered collision scenarios. The drawn trajectories are sampled using a constant velocity model and an Extended Kalman Filter. Gray ones collide, while the green ones are collision free. The black trajectory denotes the mean. The ellipses over the obstacle vehicle illustrates the positional variance.}
\label{fig:scene1}
\end{figure*}
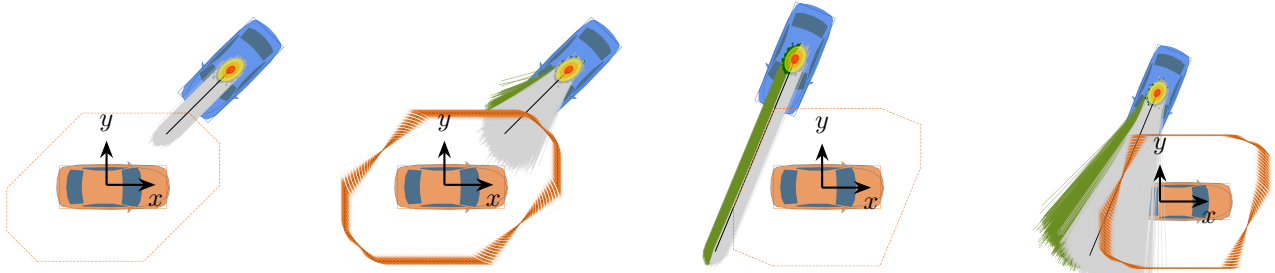
The inner integral thereby yields the expected inward normal velocity along the boundary and is given by the truncated mean of the distribution cropped at $v_n=0$, which for a Gaussian has an analytic solution,
\begin{align}
    \mathbb{E}\left[v_n^- \mid \mathbf{x}\right] &= - \mu_{v_n\mid \mathbf{x}} \, F\bigl(-z\bigr) + \sigma_{v_n\mid \mathbf{x}} \, p\bigl(z\bigr),
    \\
    &\text{with} \nonumber
\\
&z = \frac{\mu_{v_n\mid \mathbf{x}}}{\sigma_{v_n\mid \mathbf{x}}},\nonumber \\
&\mu_{v_n\mid \mathbf{x}} = \mathbf{n}^T\bigl(\mathbf{\mu}_v + \mathbf{\Sigma}_{vx} \mathbf{\Sigma}_x^{-1} (\mathbf{x}-\mathbf{\mu_x})\bigr),\nonumber \\
&\sigma_{v_n\mid \mathbf{x}}^2 = \mathbf{n}^T\bigl(\mathbf{\Sigma}_v - \mathbf{\Sigma}_{vx} \mathbf{\Sigma}_x^{-1} \mathbf{\Sigma}_{xv}\bigr)\mathbf{n}.\nonumber
\end{align}
As for the overlap probability, we can decompose the polygonal collision volume into $N$ linear edges, parametrized as $\mathbf{x}_n(s) = (1-s)\mathbf{x}_{l,i} + s\mathbf{x}_{h,i}$. We can thus rewrite the boundary integral as the sum of one-dimensional integrals along the edges,
\begin{align}
    \dot P_{bc} =& -\sum_{i=1}^N \int\limits_{s=0}^1 \mathbb{E}\left[v_n^- \mid \mathbf{x}_i(s)\right] p\bigl(\mathbf{x}_i(s)\bigr)\, \ell_i ds,
\end{align}
with the edge length $\ell_i$ as the Jacobian for the variable change. This expression can again be efficiently solved using numerical integration schemes. 
For the boundary crossing rate, the whitening of the distributions does not yield any benefits, since it relies on the evaluation of 1D-line integrals, rather than volume integrals used in the spatial overlap calculation, and is therefore not performed in this method.
For the integration of orientation uncertainties, one can proceed as in \cref{ssec:PSO} and first condition the probability rate on the orientation $\dot P_{bc}(\mathbf{x}\mid\theta)$, followed by an integration along the third dimension of the orientation-dependent polyhedron.

\section{Experiments \& Results}
In this section, we first prove the functionality of the proposed methods using Monte Carlo simulation, followed by an application study as trajectory evaluation in a sampling-based motion planner. Numerical integrations are done with Gauss-Legendre quadrature with 51 interpolation points.

\subsection{Comparison with Monte Carlo Sampling}
We consider two scenarios with a prediction horizon of \SI{3}{\second} each and simulate both with a time discretization of $\Delta t= \SI{0.1}{\second}$ once with orientation uncertainty and once without (see \cref{fig:scene1}). 
In the first scenario, the obstacle vehicle approaches the ego \ac{av} from the front left with a constant velocity, yielding a collision probability of nearly \SI{100}{\percent}. 
In the second scenario, the other vehicle approaches the ego \ac{av} with a slight offset from the side at higher velocities.  
This scenario results in a collision probability of approximately \SI{60}{\percent}. 
The exact simulation data are shown in Table \ref{tab:data}.

We first evaluate the collision probability using five different approaches. For the evaluation of the spatial overlap probability, we do Monte Carlo Sampling for each state along the predicted trajectory independently and calculate the proportion of collided samples per timestep (MC State Sampling). 
Second, we calculate the probability of spatial overlap for each time step using \cref{eq:p_ov}, and third, we evaluate the collision probability by assuming temporal independence as described by \cref{eq:ind}.
To compare the probability of boundary penetrations (\cref{eq:int_bc}), we sample complete trajectories using Monte Carlo (MC Traj. Sampling). A trajectory contributes to the collision probability only after the time step of collision. 
In both sampling-based methods, we sample 25000 states and trajectories, respectively. 
In \cref{fig:1}, the resulting collision probabilities are plotted over time for both scenarios when the orientation uncertainties are neglected.
\begin{figure}[t]
    \centering
    \vspace*{2mm}
\input{figures/prob_plot}
\caption{Collision probability without orientation uncertainties }
\label{fig:1}
\end{figure}
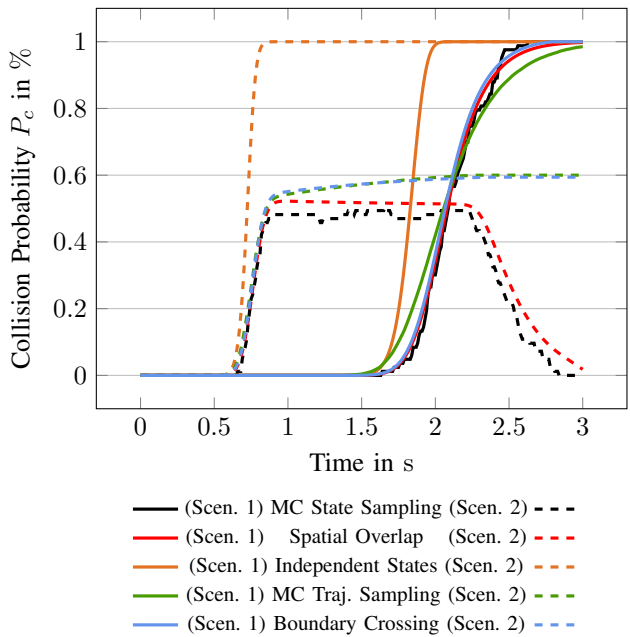
\begin{table}[!bt]
\centering
\caption{Initial mean and variance of the simulated ego and obstacles, each in the global coordinate system.}
\label{tab:data}
\begin{tabular}{lccc}
\toprule
Value & Ego \ac{av} & Obs. scenario 1 & Obs. scenario 2\\
\midrule
$\mu_\mathbf{x}^0$ in \si{\meter} & $(0,0)$ & $(5.5,5.5)$ & $(-1.5,6)$ \\
$\Sigma_\mathbf{x}^0$ in \si{\meter\squared}  
& $\setlength{\arraycolsep}{1pt} \begin{bmatrix} 0.1^2& 0 \\ 0 & 0.05^2\end{bmatrix} $ 
& $\setlength{\arraycolsep}{1pt} \begin{bmatrix} 0.2^2& 0 \\ 0 & 0.1^2\end{bmatrix} $ 
& $\setlength{\arraycolsep}{1pt} \begin{bmatrix} 0.2^2& 0 \\ 0 & 0.1^2\end{bmatrix} $\\
$\mu_v^0$ in \si{\meter\per\second} & $0$ &$ 1.4$ & $3.25$ \\
$\sigma_v^{0}$ in \si{\meter\per\second} &  $0.1$ & $0.1$ & $0.1$ \\
$\theta^0$ in \si{\radian}& $0$ & $-\pi/4$ & $-3\pi/8$ \\
$\sigma_\theta^0$ in \si{\radian}& $0/0.1$ & $0/0.1$ & $0/0.1$ \\
\bottomrule
\end{tabular}
\end{table}
For the colliding scenario (scenario 1), all approaches predict a collision probability of 1. However, it can be observed that the assumption of independent states does not correspond to reality, resulting in an excessively conservative collision probability, which is more than \SI{0.5}{\second} earlier at \SI{100}{\percent} than the other methods.
All other approaches yield similar results with an increasing collision probability between \SI{1.5}{\second} and \SI{2.5}{\second}. Only the trajectory sampling approach stands out to some extent as it ascends earlier, but with a slightly lower slope.
In the second scenario in \cref{fig:1}, where a collision probability of approximately \SI{60}{\percent} can be observed, the independence assumption again significantly overestimates the collision risk.
Spatial overlap yields a maximum collision probability of \SI{53}{\percent} after ca. \SI{0.8}{\second} and a decreasing overlap probability towards the end of the time period, when the obstacle vehicle has already passed the rear of the ego \ac{av}. The tendency matches state sampling, but predicts a minimal higher maximum collision probability.
In contrast, the boundary crossing probability initially rises similarly to the overlap probability, but continues to increase when the obstacle vehicle drives along the collision volume. 
\begin{figure}[t]
    \centering
    \vspace*{2mm}
\input{figures/prob_plot_with_orient}
\caption{Collision probability with orientation uncertainties }
\label{fig:2}
\end{figure}
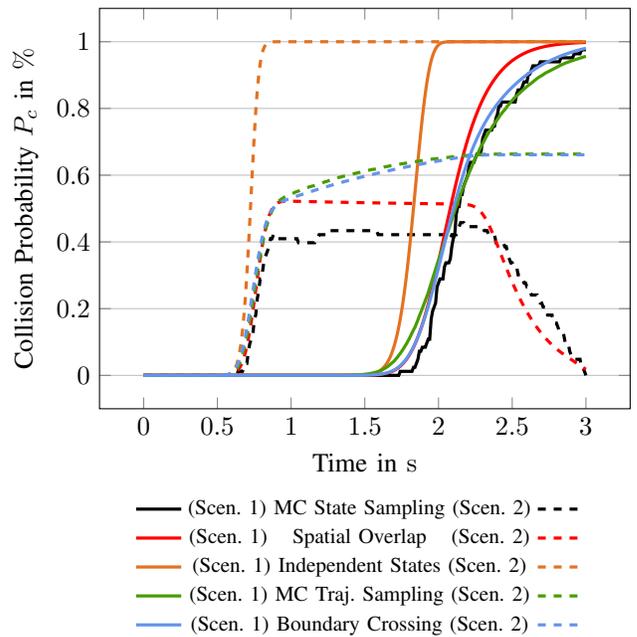
It correctly matches the collision probability of the trajectory sampling approach and is thus able to correctly catch the temporal dependencies due to the probabilistic normal velocity along the boundary. 

In \cref{fig:2}, the collision probabilities for the scenarios including orientation uncertainties are illustrated.
The overall tendencies are identical to the first case. The spatial overlap calculation again overestimates the probability by approximately \SI{10}{\percent} for the second scenario, while the boundary crossing probability aligns well with the collision probabilities of the sampled trajectories.
However, when including the orientation in the analysis, the overall collision probability in the second scenario increases from \SI{60}{\percent} to \SI{66}{\percent}, highlighting the necessity to include orientation deviations of predictions in the collision analysis, since small changes in the initial orientation can result in significantly different collision probabilities. 

Next, we compare the runtime of our proposed methods. For the application for trajectory evaluation in motion planning in autonomous driving, it is important to achieve fast calculation times. In \cref{tab:times}, we list the mean computation times on a 12th Gen Intel Core i7 processor.
Both proposed methods are significantly faster than the sampling-based approaches.  The computation times for our overlap probability implementation are lower than those for the border crossing rate. However, both remain below \SI{0.1}{\second} and \SI{0.01}{\second}, with and without orientation, respectively.
\begin{table}[!hbt]
\centering
\caption{Runtime Analysis of the collision probability along the complete trajectory in \si{\second}.}
\label{tab:times}
\begin{tabular}{lcc}
\toprule
Method & Without Orientation & With Orientation\\
\midrule
MC State Sampling &0.3 &10.3 \\
Spatial Overlap & 0.001& 0.02\\
MC Traj. Sampling & 1.9 & 2.5\\
Boundary Crossing & 0.003& 0.09\\
\bottomrule
\end{tabular}
\end{table}
\newline

\subsection{Application Example in an Analytic Motion Planner}
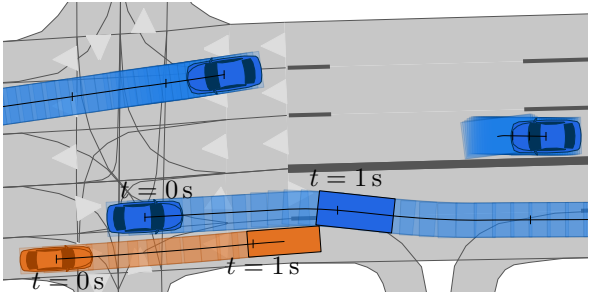
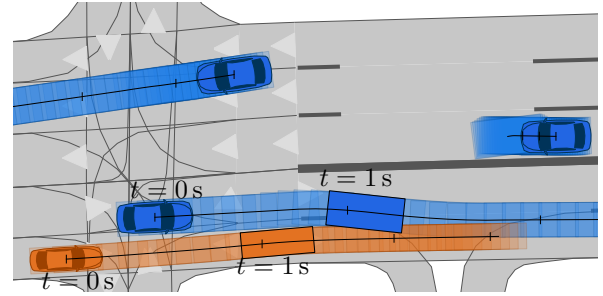
\begin{figure*}[!t]
\centering
\subfloat[][Planning without residual risk. Potentially colliding trajectories are filtered out, resulting in an end of the simulation after \SI{1.2}{\second} since no collision-free trajectories are found.]{\label{fig:sim1}
\input{figures/sim2}}
\hfill
\subfloat[][Planning with probability based collision estimation. The Planner can select trajectories with residual risk.]{\label{fig:sim2}
\input{figures/sim}}
\caption{Simulated cut-in scenario. The ego \ac{av} is the orange vehicle, while all blue ones are controlled by the simulation. The occupancies show the final trajectories. The highlighted occupancy illustrates in both images the final time step of the conservative approach after \SI{1.2}{\second}. }
\label{fig:sim}
\end{figure*}

Finally, we integrate our proposed collision probability calculations into a sampling-based motion planning framework and simulate a critical cut in maneuver. In each iteration, the planner evaluates 600 sampled trajectories. The predicted collision risk is thereby incorporated into the cost function to select the trajectory with the lowest risk. As a comparison, we use a conservative collision check, which filters out all trajectories intersecting the oriented bounding box that covers the $3 \sigma $ area around the prediction of an obstacle, avoiding any risk in the planning process. 
The uncertainties are forecasted along the obstacle trajectory using a Kalman Filter assuming a symmetric pose variance of $\sigma_x =\sigma_y= \SI{0.3}{\meter}$ and a velocity variation of $\sigma_v = \SI{0.15}{\meter\per\second}$. 
In \cref{fig:sim}, we show the resulting final paths. 
For the situation where all collision risk is avoided (\cref{fig:sim1}), the planner is only able to plan for \SI{1.2}{\second}. Afterwards, it does not find any collision-free trajectory. In contrast, by allowing some residual risk, the planner is able to solve the situation by selecting a potentially riskier trajectory, which decelerates the car strongly enough to eventually avoid the rear-end collision (\cref{fig:sim2}). The same results can be obtained using both proposed methods, the maximum spatial overlap and the boundary crossing probability.

Since the implementation of our proposed collision prediction formulations can easily be vectorized, the computation time per trajectory batch remains small with approximately \SI{0.08}{\second} for the overlap probability and \SI{0.15}{\second} for the boundary penetrations.

\section{Discussion}
Our results demonstrate the effectiveness of both proposed methods for collision risk prediction.
The analytic formulation of the spatial overlap probability correctly reflects the overlap probability calculated via Monte Carlo Sampling, while the boundary crossing probability aligns well with the collision risk estimation by sampling complete trajectories. We also proved that assuming independence of states along a trajectory is not a valid assumption for autonomous driving. 
While the collision analysis based on spatial overlap can efficiently be used as a collision metric, it does not correctly present the underlying ground truth collision risk apparent in motion planning, where trajectories need to be assessed with respect to the safety of collisions through body intrusions into other vehicles. In contrast, our second approach can be used to predict the collision risk more accurately as it estimates the collision probability based on the stochasticity of boundary crossings; however, at minimal higher computation costs.

The experiments also allow the conclusion that the inclusion of orientational variance into the collision risk estimation may be necessary in order to capture more accurately the additional collision risk induced by small errors in the detected initial heading of obstacles. We showed in an exemplary edge case scenario that the collision probability can increase significantly by \SI{10}{\percent} from \SIrange{60}{66}{\percent}.

\section{Conclusion and Future Outlook}
This paper presents two efficient methods for the accurate prediction of the collision risk along planned trajectories for autonomous driving. Both methods are based on the assumption of prevailing state uncertainties caused by sensor and prediction inaccuracies. A fast estimate can be obtained by computing the probability of spatial overlap, while a more accurate method based on boundary crossing rates yields more precise results. The findings also highlight the necessity to include obstacle orientations in the collision risk calculation, since already small deviations in the detected heading can yield notable differences in the estimated collision probability. 

Future research will extend the framework beyond Gaussian assumptions to better capture real-world sensor noise and prediction errors. In addition, both methods will be tested under real-world driving conditions to further validate robustness and efficiency.

%% file: figures/collision_octagon.tex

\begin{tikzpicture}[>={Stealth}]
      \node[anchor=south west, inner sep=0] (img) at (0,0) {\includegraphics[width=0.85\columnwidth, trim=2.2cm 2.5cm 2cm 3cm, clip]{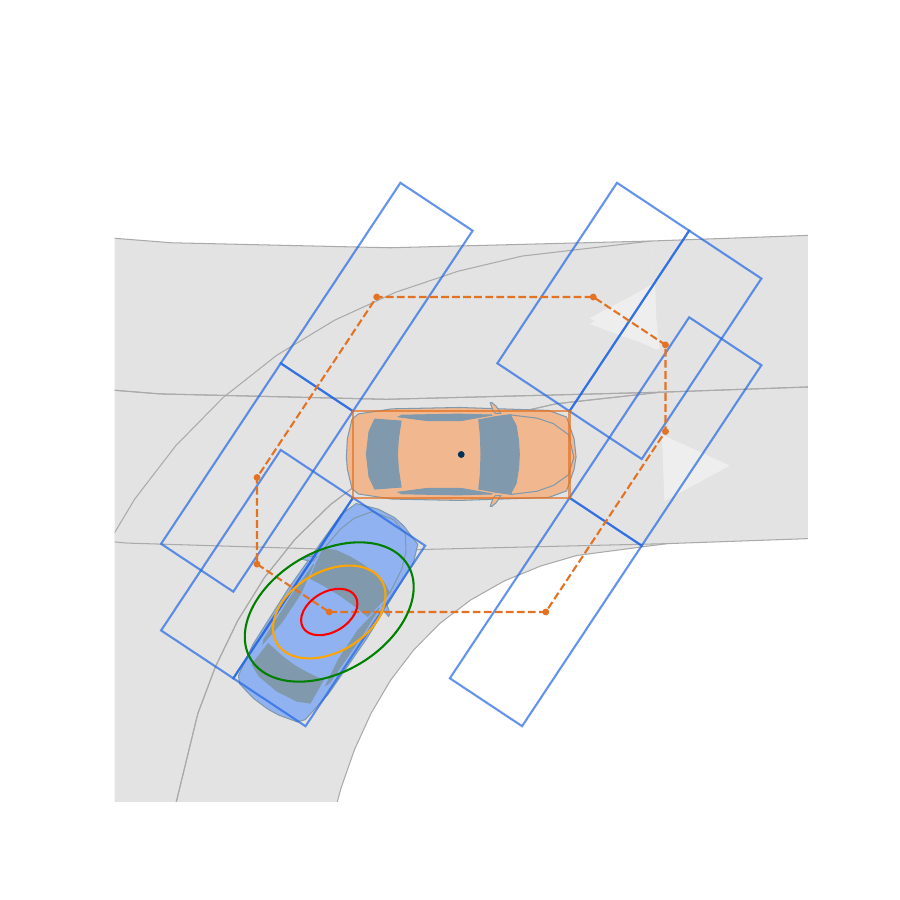}};

\begin{scope}[x={(img.south east)},y={(img.north west)}]
    \draw[->, thick] (0.508,0.52) -- (0.508,0.65) node[above] {$y$};
    \draw[->, thick] (0.508,0.52) -- (0.62,0.52) node[right] {$x$};

    \node at (0.3, 0.2) {1};
    \node at (0.64, 0.2) {2};
    \node at (0.86, 0.56) {3};
    \node at (0.86, 0.72) {4};
    \node at (0.71, 0.84) {5};
    \node at (0.37, 0.84) {6};
    \node at (0.15, 0.48) {7};
    \node at (0.15, 0.32) {8};

    \draw[->, thick, orange] (0.35,0.2)   -- ++(0:0.25);
    \draw[<-, thick, orange] (0.42,0.84)   -- ++(0:0.25);
     
     \draw[->, thick, orange] (0.71,0.28) -- ++(59.7:0.25);
     \draw[<-, thick, orange] (0.19,0.56) -- ++(59.7:0.25);

     \draw[->, thick, orange] (0.86,0.6) -- ++(90:0.08);
     \draw[<-, thick, orange] (0.15,0.36) -- ++(90:0.08);

     \draw[->, thick, orange] (0.82,0.76) -- ++(90 +55:0.08);
     \draw[<-, thick, orange] (0.25,0.23) -- ++(90 +55:0.08);
    
\end{scope}
\end{tikzpicture}


%% file: figures/polyhedron.tex
\begin{tikzpicture}[>={Stealth}] 

\node[ anchor=south west,inner sep=0, yscale=0.92] (img) at (0,0) {\includegraphics[width=0.55\columnwidth, trim=4.5cm 4.3cm 7.1cm 4cm, clip]{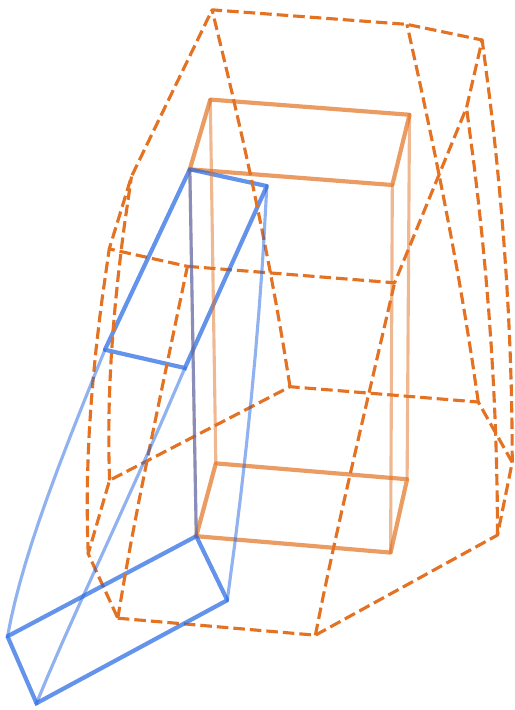}};

\begin{scope}[x={(img.south east)},y={(img.north west)}]

    \draw[->, thick,shift={(0.2,0.05)}](0.38,0.24) -- (0.41,0.36) node[right] {$y$};
    \draw[->, thick, shift={(0.2,0.05)}] (0.38,0.24) -- (0.54,0.23) node[below ] {$x$};
    \draw[->, thick,shift={(0.2,0.05)}](0.38,0.24) -- (0.38,0.38) node[above] {$\theta$};

    \begin{scope}[shift={(1.25,0.6)},rotate=-90, xscale=0.14, yscale=0.4] 
         \draw[domain=-3:3,smooth,variable=\x,TUMBlue,thick] plot ({\x},{1/sqrt(2*pi)*exp(-(\x)^2/2)});
        \node[] at (0,0.55) {$p_\theta$};
    \draw[->, thick] (3.2,0) -- (-3.7,0) node[right] {$\theta$};

    \foreach \x in {-3,3} {
        \draw (-\x,0.02) -- (-\x,-0.02) node[left] {$\x\, \sigma_\theta$};
        
    }
    \draw (0,0.02) -- (0,-0.02) node[left] {$ \mu_\theta$};
    \draw[gray,dashed] (0,0) -- (0,{1/sqrt(2*pi)*exp(-(0)^2/2)});
    \end{scope}
\end{scope}
\end{tikzpicture}

%% file: figures/whitened_oct.tex
\begin{tikzpicture}[>={Stealth}, scale=0.9]
  \node[anchor=south west, inner sep=0,] (img) at (0,0) {\includegraphics[width=0.7\columnwidth, trim=5.5cm 4cm 2.1cm 2cm, clip]{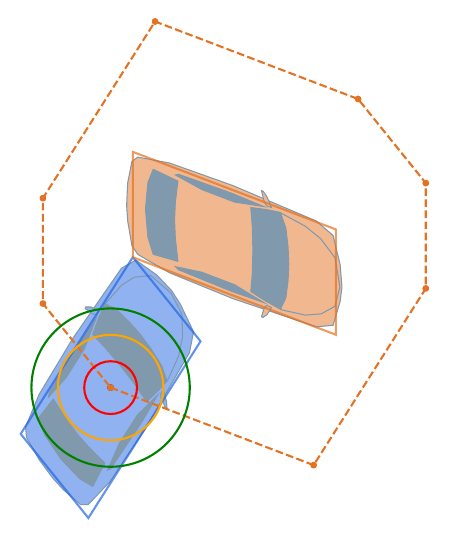}};

\begin{scope}[x={(img.south east)},y={(img.north west)}]

    \draw[->, thick] (0.245,0.29) -- (0.245,0.47) node[above] {$\hat y$};
    \draw[->, thick] (0.245,0.29) -- (0.45,0.29) node[below ] {$\hat x$};

    \node at (0.22, 0.27) {1};
    \node at (0.695, 0.11) {2};
    \node at (0.97, 0.47) {3};
    \node at (0.97, 0.67) {4};
    \node at (0.805, 0.85) {5};
    \node at (0.345, 1) {6};
    \node at (0.06, 0.64) {7};
    \node at (0.06, 0.44) {8};
    
\end{scope}
\end{tikzpicture}

%% file: figures/boundary_cross.tex
\begin{tikzpicture}[>={Stealth}]
  \node[anchor=south west, inner sep=0] (img) at (0,0) 
  {\includegraphics[width=0.85\columnwidth, trim=2.5cm 1.8cm 3.8cm 4.3cm, clip]{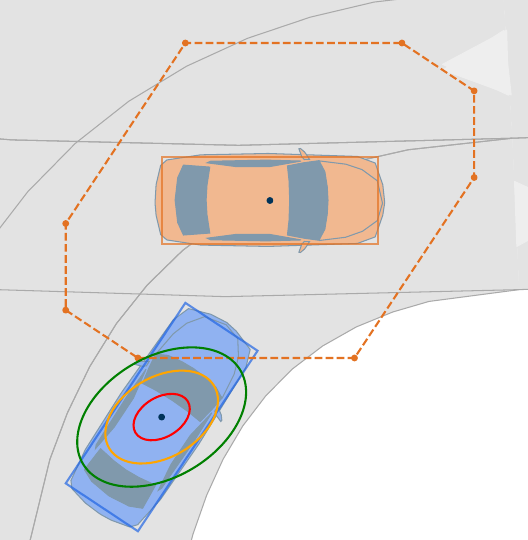}};

\begin{scope}[x={(img.south east)},y={(img.north west)}]

    \draw[->, thick] (0.512,0.625) -- (0.512,0.76) node[above] {$y$};
    \draw[->, thick] (0.51,0.625) -- (0.64,0.625) node[right] {$x$};

    \draw[->,  very thick] (0.305,0.225) -- (0.45,0.45) node[right] {$\mathbf{v}$};
    \draw[->,  thick] (0.305,0.225) -- (0.305,0.45) node[above] {$\mathbf{v}_n$};
    \draw[->,  thick] (0.305,0.225) -- (0.45,0.225) node[right] {$\mathbf{v}_t$};
    \draw[->,  thick, draw=myorange] (0.55,0.34) -- (0.55,0.27) node[right] {$\mathbf{n}$};
    \node at (0.245, 0.30) {1};
    \node at (0.675, 0.30) {2};
    \node at (0.94, 0.67) {3};
    \node at (0.94, 0.83) {4};
    \node at (0.76, 0.97) {5};
    \node at (0.35, 0.97) {6};
    \node at (0.08, 0.59) {7};
    \node at (0.08, 0.425) {8};
   
\end{scope}
\end{tikzpicture}

%% file: figures/scene1.tex
\begin{tikzpicture}[>={Stealth}]
  \node[anchor=south west, inner sep=0] (img) at (0,0) {\includegraphics[width=0.47\columnwidth, trim=1cm 3.2cm 0.7cm 2.5cm, clip]{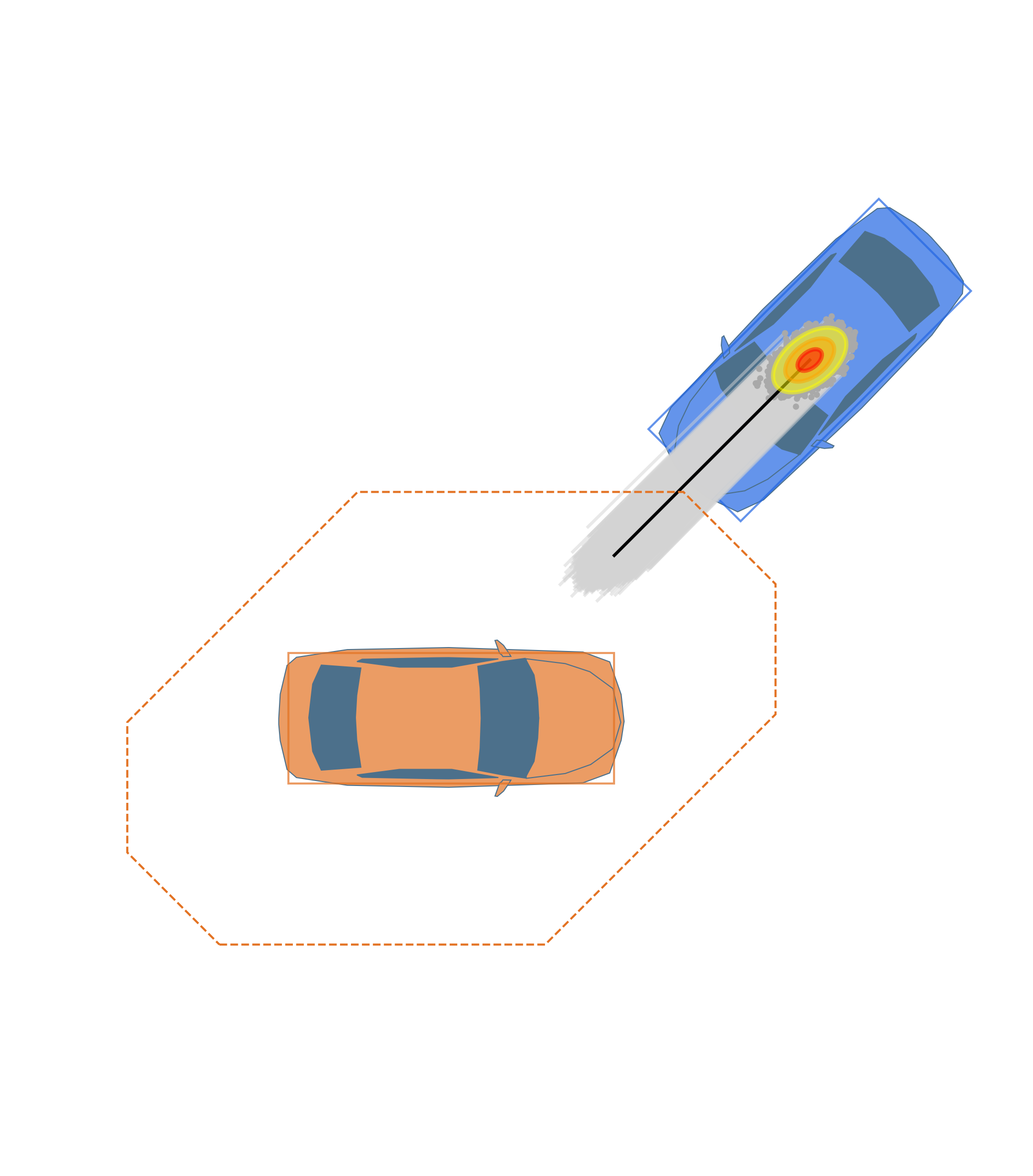}};

\begin{scope}[x={(img.south east)},y={(img.north west)}]

    \draw[->, thick](0.4,0.33) -- (0.4,0.49) node[above] {$y$};
    \draw[->, thick, ] (0.4,0.33) -- (0.56,0.33) node[below ] {$x$};
    
\end{scope}
\end{tikzpicture}

%% file: figures/scene1_orient.tex
\begin{tikzpicture}[>={Stealth}]
  \node[anchor=south west, inner sep=0] (img) at (0,0) {\includegraphics[width=0.47\columnwidth, trim=1cm 3.2cm 0.7cm 2.5cm, clip]{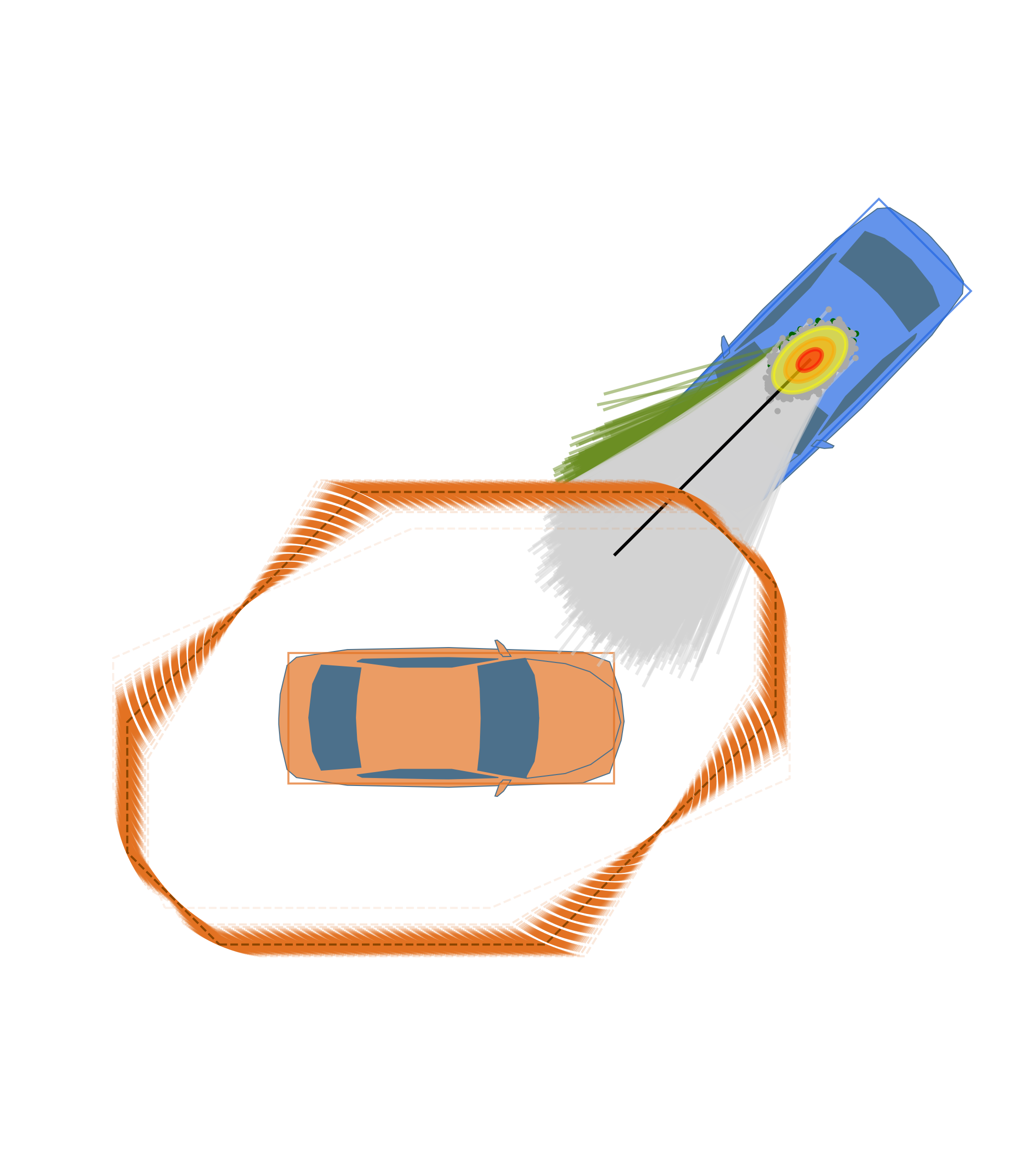}};

\begin{scope}[x={(img.south east)},y={(img.north west)}]
    
    \draw[->, thick](0.4,0.33) -- (0.4, 0.49) node[above] {$y$};
    \draw[->, thick, ] (0.4,0.33) -- (0.56,0.33) node[below ] {$x$};
    
\end{scope}
\end{tikzpicture}

%% file: figures/scene2.tex
\begin{tikzpicture}[>={Stealth}]
  \node[anchor=south west, inner sep=0] (img) at (0,0) {\includegraphics[width=0.47\columnwidth, trim=-1cm 3.2cm 2.7cm 2.5cm, clip]{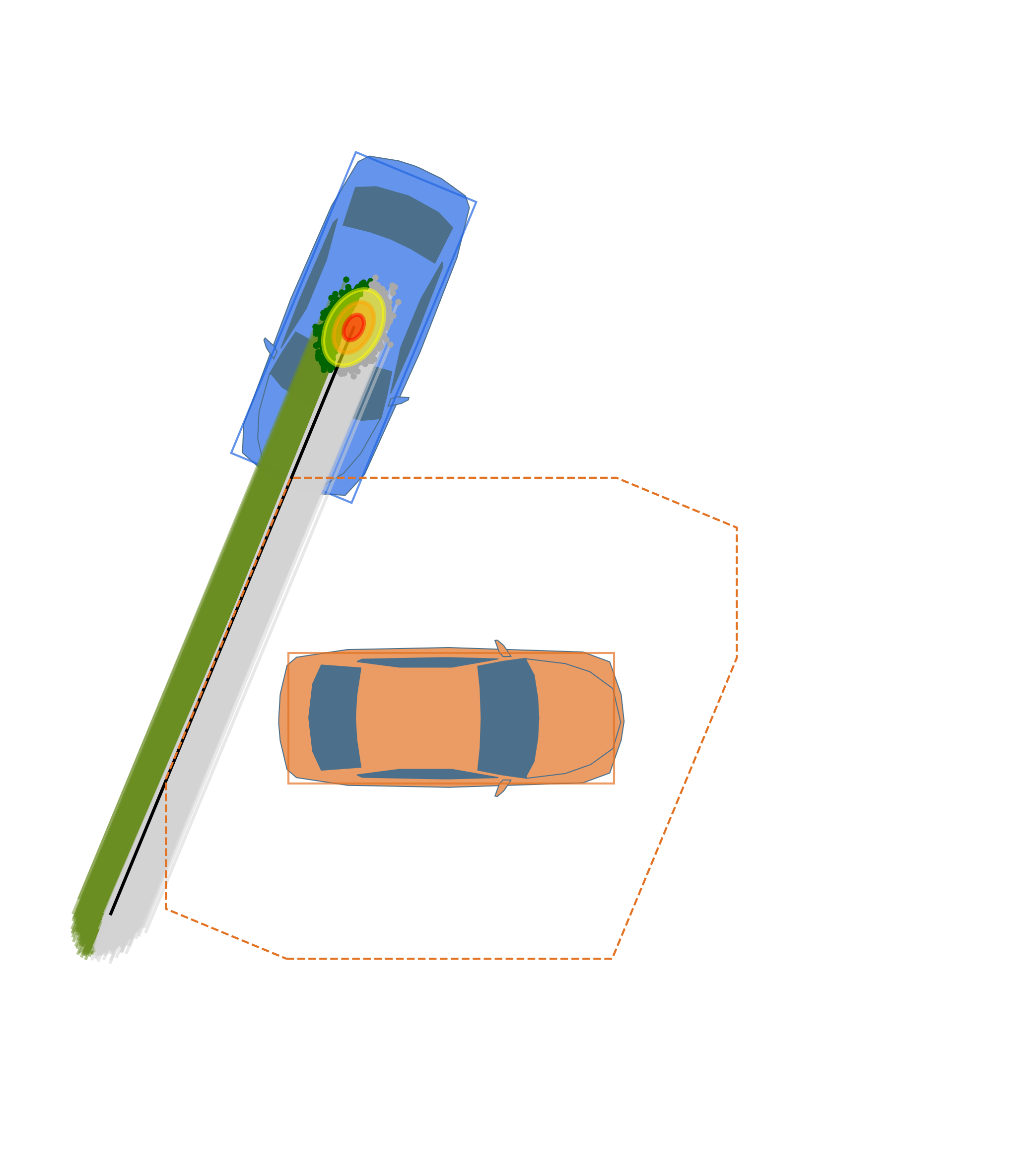}};

\begin{scope}[x={(img.south east)},y={(img.north west)}]
    
    \draw[->, thick](0.53,0.32) -- (0.53,0.48) node[above] {$y$};
    \draw[->, thick, ] (0.53,0.32) -- (0.69,0.32) node[below ] {$x$};
    
\end{scope}
\end{tikzpicture}

%% file: figures/scene2_orient.tex
\begin{tikzpicture}[>={Stealth}]
  \node[anchor=south west, inner sep=0] (img) at (0,0) {\includegraphics[width=0.47\columnwidth, trim=-1cm 3.2cm 2.7cm 2.5cm, clip]{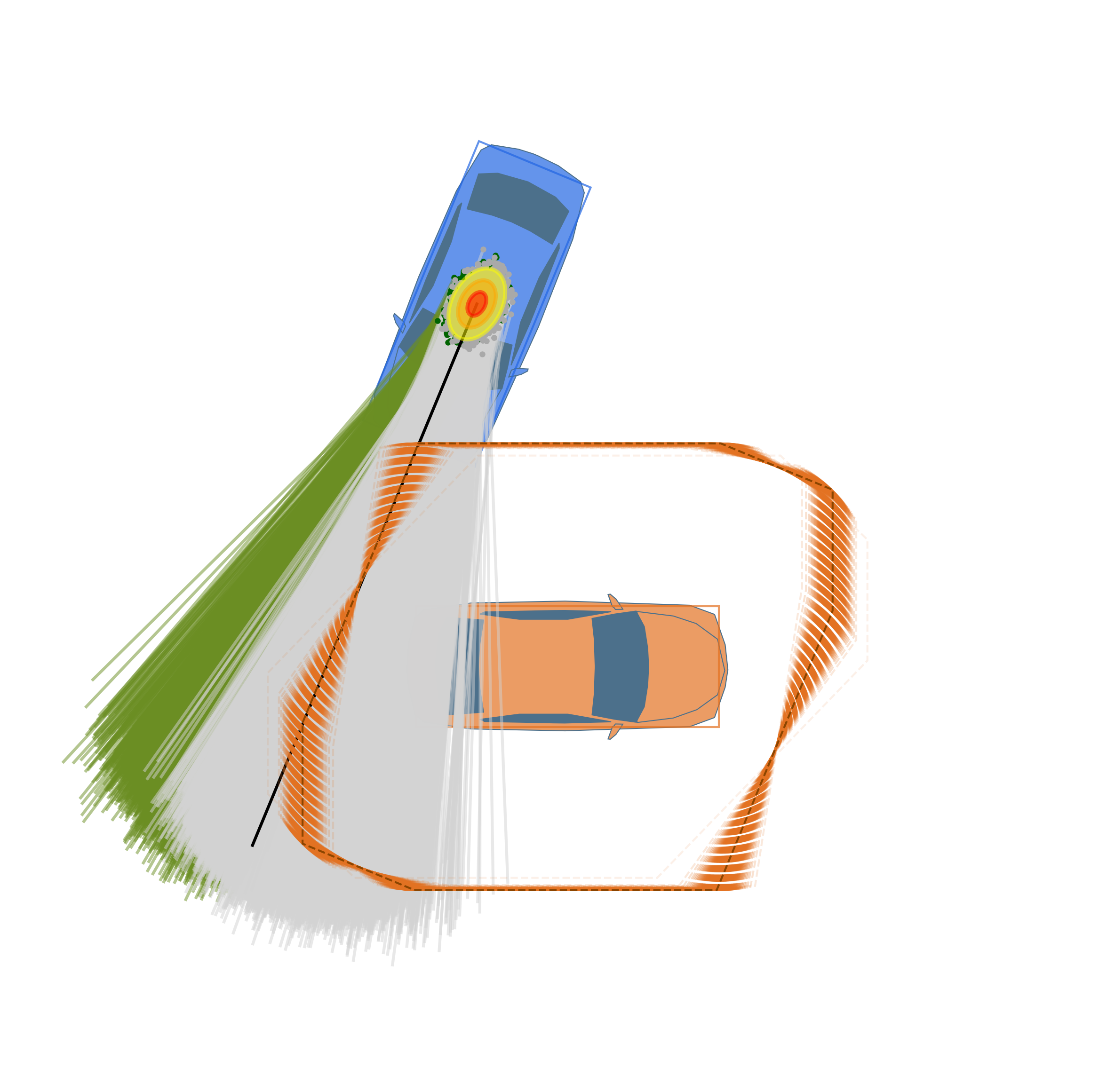}};

\begin{scope}[x={(img.south east)},y={(img.north west)}]
    
    \draw[->, thick](0.53,0.32) -- (0.53,0.48) node[above] {$y$};
    \draw[->, thick, ] (0.53,0.32) -- (0.69,0.32) node[below ] {$x$};
    
\end{scope}
\end{tikzpicture}

%% file: figures/prob_plot.tex
\begin{tikzpicture}[>={Stealth}]

\begin{axis}[
    width=\columnwidth,
    height = 0.8*\columnwidth,
    xlabel={Time in \si{\second}},
    ylabel={Collision Probability $P_c$ in \si{\percent}},
    ymajorgrids=true,
    legend style={at={(0.5,-0.2)}, 
        anchor=north,
        legend columns=2,
        draw=none,          
        font=\footnotesize,
        },
     legend cell align={center},
]
\pgfmathsetmacro{\dt}{0.01}

obs 0 wo orient: 
\addplot+[
very thick,
color = TUMBlack,
mark=none
] table [x expr=\dt*\coordindex, y=MC_state_rect_wo_orientation_0, col sep=comma] {figures/0_data_prob_rates.csv};
\addlegendentry{(Scen. 1) MC State Sampling (Scen. 2)}
\addplot+[
very thick,
color = TUMBlack,
dashed,
mark=none
] table [x expr=\dt*\coordindex, y=MC_state_rect_wo_orientation_1, col sep=comma] {figures/0_data_prob_rates.csv};
\addlegendentry{ }

\addplot+[
very thick,
color = TUMRed,
mark=none,
] table [x expr=\dt*\coordindex, y=State_Maha_wo_orientation_0, col sep=comma] {figures/0_data_prob_rates.csv};
\addlegendentry{(Scen. 1)\;\;\; Spatial Overlap\;\;\;  (Scen. 2)}

\addplot+[
very thick,
color = TUMRed,
dashed,
mark=none,
] table [x expr=\dt*\coordindex, y=State_Maha_wo_orientation_1, col sep=comma] {figures/0_data_prob_rates.csv};
\addlegendentry{}

\addplot[
very thick,
color = TUMOrange,
mark=none,
] table [x expr=\dt*\coordindex, y=State_Maha_wo_orientation_0, col sep=comma] {figures/0_data_prob_cum.csv};
\addlegendentry{(Scen. 1) Independent States (Scen. 2)}
\addplot[
very thick,
color = TUMOrange,
dashed,
mark=none,
] table [x expr=\dt*\coordindex, y=State_Maha_wo_orientation_1, col sep=comma] {figures/0_data_prob_cum.csv};
\addlegendentry{}

\addplot[
very thick,
color = TUMGreen,
mark=none
] table [x expr=\dt*\coordindex, y=MC_CV_wo_orientation_0, col sep=comma] {figures/0_data_prob_cum.csv};
\addlegendentry{(Scen. 1) MC Traj. Sampling (Scen. 2)}
\addplot[
very thick,
color = TUMGreen,
dashed,
mark=none
] table [x expr=\dt*\coordindex, y=MC_CV_wo_orientation_1, col sep=comma] {figures/0_data_prob_cum.csv};
\addlegendentry{ }
\addplot[
very thick,
color = TUMBlue,
mark=none,
] table [x expr=\dt*\coordindex, y=CEP_wo_orientation_0, col sep=comma] {figures/0_data_prob_cum.csv};
\addlegendentry{(Scen. 1) Boundary Crossing (Scen. 2)}
\addplot[
very thick,
color = TUMBlue,
mark=none,
dashed,
] table [x expr=\dt*\coordindex, y=CEP_wo_orientation_1, col sep=comma] {figures/0_data_prob_cum.csv};
\addlegendentry{}

\end{axis}
\end{tikzpicture}

%% file: figures/prob_plot_with_orient.tex
\begin{tikzpicture}[>={Stealth}]

\begin{axis}[
    width=\columnwidth,
    height = 0.8*\columnwidth,
    xlabel={Time in \si{\second}},
    ylabel={Collision Probability $P_c$ in \si{\percent}},
    ymajorgrids=true,
    legend style={at={(0.5,-0.2)}, 
        anchor=north,
        legend columns=2,
        draw=none,          
        font=\footnotesize,
        },
     legend cell align={center},
]
\pgfmathsetmacro{\dt}{0.01}

\addplot+[
very thick,
color = TUMBlack,
mark=none
] table [x expr=\dt*\coordindex, y=MC_state_rect_with_orientation_0, col sep=comma] {figures/0_data_prob_rates.csv};
\addlegendentry{(Scen. 1) MC State Sampling (Scen. 2)}
\addplot+[
very thick,
color = TUMBlack,
dashed,
mark=none
] table [x expr=\dt*\coordindex, y=MC_state_rect_with_orientation_1, col sep=comma] {figures/0_data_prob_rates.csv};
\addlegendentry{ }

\addplot+[
very thick,
color = TUMRed,
mark=none,
] table [x expr=\dt*\coordindex, y=State_Maha_with_orientation_0, col sep=comma] {figures/0_data_prob_rates.csv};
\addlegendentry{(Scen. 1)\;\;\; Spatial Overlap\;\;\;  (Scen. 2)}

\addplot+[
very thick,
color = TUMRed,
dashed,
mark=none,
] table [x expr=\dt*\coordindex, y=State_Maha_with_orientation_1, col sep=comma] {figures/0_data_prob_rates.csv};
\addlegendentry{}

\addplot[
very thick,
color = TUMOrange,
mark=none,
] table [x expr=\dt*\coordindex, y=State_Maha_with_orientation_0, col sep=comma] {figures/0_data_prob_cum.csv};
\addlegendentry{(Scen. 1) Independent States (Scen. 2)}
\addplot[
very thick,
color = TUMOrange,
dashed,
mark=none,
] table [x expr=\dt*\coordindex, y=State_Maha_with_orientation_1, col sep=comma] {figures/0_data_prob_cum.csv};
\addlegendentry{}

\addplot[
very thick,
color = TUMGreen,
mark=none
] table [x expr=\dt*\coordindex, y=MC_CV_with_orientation_0, col sep=comma] {figures/0_data_prob_cum.csv};
\addlegendentry{(Scen. 1) MC Traj. Sampling (Scen. 2)}
\addplot[
very thick,
color = TUMGreen,
dashed,
mark=none
] table [x expr=\dt*\coordindex, y=MC_CV_with_orientation_1, col sep=comma] {figures/0_data_prob_cum.csv};
\addlegendentry{ }
\addplot[
very thick,
color = TUMBlue,
mark=none,
] table [x expr=\dt*\coordindex, y=CEP_with_orientation_0, col sep=comma] {figures/0_data_prob_cum.csv};
\addlegendentry{(Scen. 1) Boundary Crossing (Scen. 2)}
\addplot[
very thick,
color = TUMBlue,
mark=none,
dashed,
] table [x expr=\dt*\coordindex, y=CEP_with_orientation_1, col sep=comma] {figures/0_data_prob_cum.csv};
\addlegendentry{}

\end{axis}
\end{tikzpicture}

%% file: figures/sim2.tex
\begin{tikzpicture}[scale=0.9]

\node[anchor=south west,inner sep=0 , rotate = 90] (image) at (0,0) {\includegraphics[width=0.45\columnwidth,trim=1.5cm 2cm 8.5cm 5.5cm, clip,]{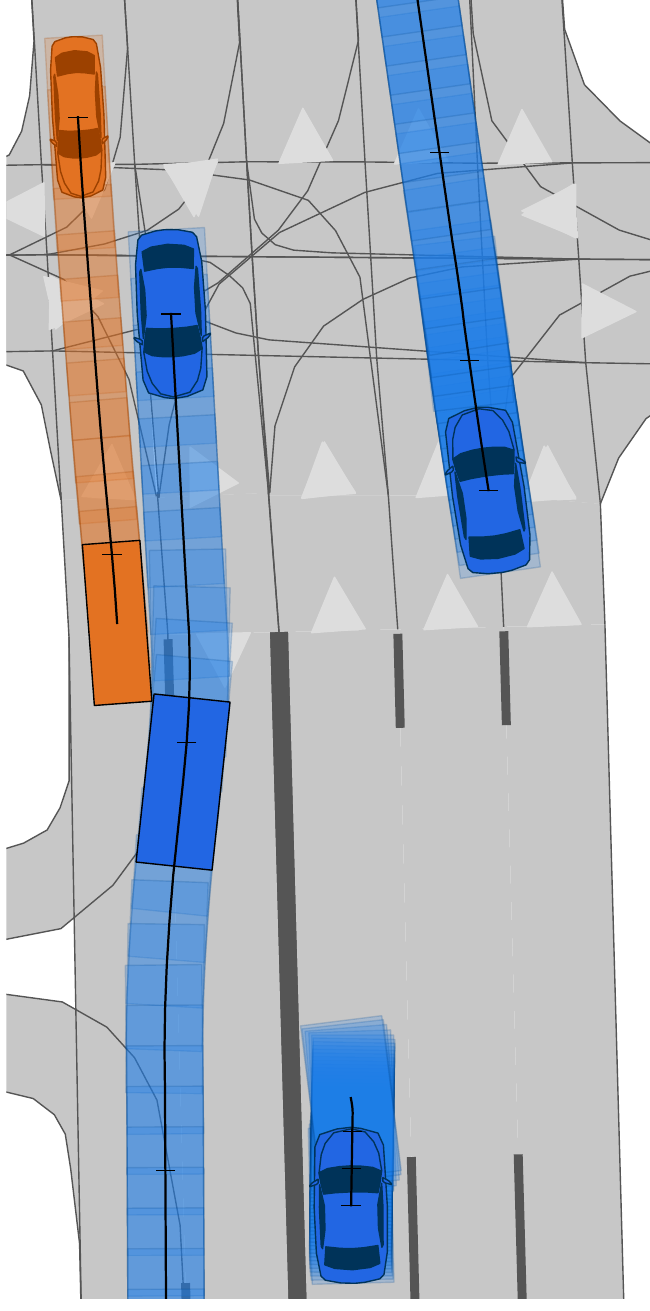}};

\begin{scope}[x={(image.south east)},y={(image.north west)}]

    \node[] at (0.05,0.89) {$t = \SI{0}{\second}$}; 
    \node[] at (0.1,0.56) {$t = \SI{1}{\second}$}; 
    \node[] at (0.36,0.74) {$t = \SI{0}{\second}$}; 
    \node[] at (0.4,0.42) {$t = \SI{1}{\second}$}; 
       
        \end{scope}
\end{tikzpicture}

%% file: figures/sim.tex
\begin{tikzpicture}[scale=0.9]
            \node[anchor=south west,inner sep=0, rotate=90, ] (image) at (0,0){\includegraphics[width=0.45\columnwidth,trim=1.5cm 2cm 8.5cm 5.5cm, clip ]{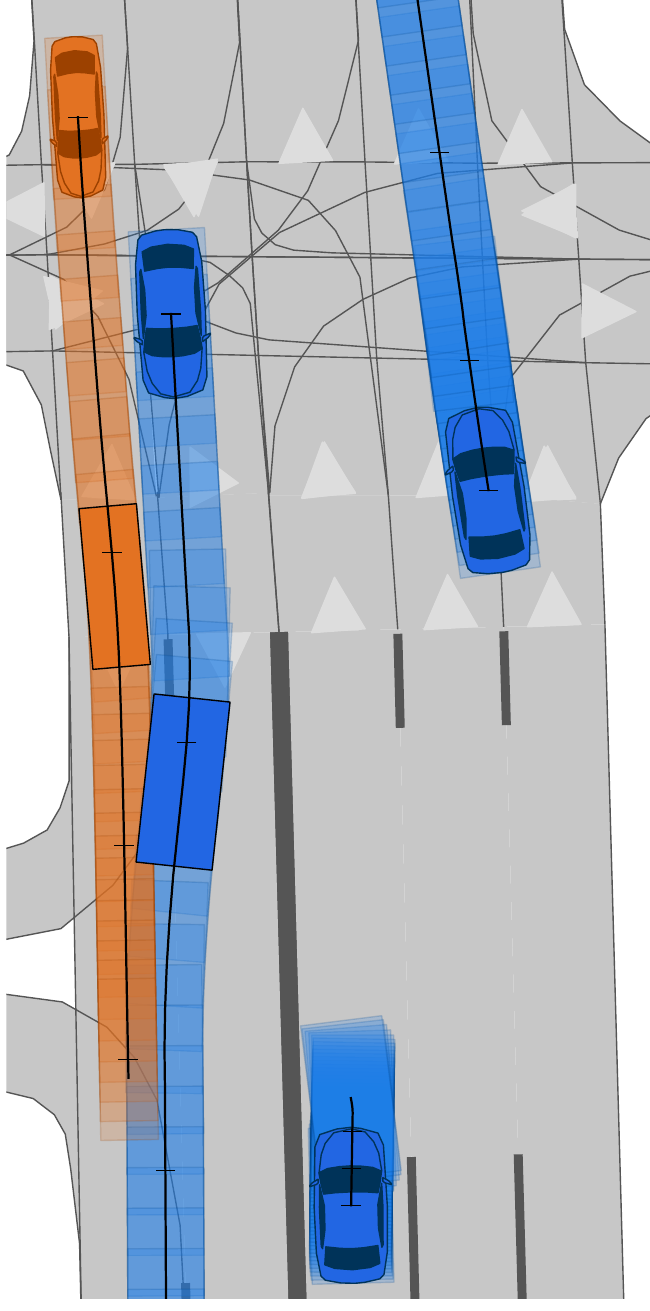}};

\begin{scope}[x={(image.south east)},y={(image.north west)}]

    \node[] at (0.05,0.89) {$t = \SI{0}{\second}$}; 
    \node[] at (0.1,0.56) {$t = \SI{1}{\second}$}; 
    \node[] at (0.36,0.74) {$t = \SI{0}{\second}$}; 
    \node[] at (0.4,0.42) {$t = \SI{1}{\second}$}; 
        \end{scope}
\end{tikzpicture}